\documentclass[11pt]{article}
\usepackage[utf8]{inputenc}        
\usepackage[T1]{fontenc}           
\usepackage{lmodern}               
\usepackage{geometry}              
\usepackage{microtype}             
\usepackage{graphicx}              
\usepackage{amsmath, amssymb, amsfonts}
\usepackage{enumitem}
\usepackage{xcolor}                
\usepackage{hyperref}              
\usepackage{cleveref}              

\usepackage{algorithm}
\usepackage{algorithmic}

\usepackage{newfloat}
\usepackage{listings}
\floatstyle{ruled}
\newfloat{listing}{tb}{lst}
\floatname{listing}{Listing}

\usepackage{acro}
\DeclareAcronym{LLM}{short = LLM, long = Large Language Model}
\DeclareAcronym{MLA-BiTe}{short = MLA-BiTe, long = MultiLingual Augmented Bias Testing}
\DeclareAcronym{LangBiTe}{short = LangBiTe, long = Language Bias Testing}
\DeclareAcronym{BLEU}{short = BLEU, long = Bilingual Evaluation Understudy}
\DeclareAcronym{NLP}{short = NLP, long = Natural Language Processing}

\usepackage[numbers]{natbib}
\title{Mind the Language Gap: Automated and Augmented Evaluation of Bias in LLMs for High- and Low-Resource Languages}

\author{
Alessio Buscemi$^{1}$, 
C\'edric Lothritz$^{1}$, 
Sergio Morales$^{2}$, 
Marcos Gomez-Vazquez$^{1}$,\\
Robert Claris\'o$^{2}$, 
Jordi Cabot$^{1,3}$, 
Germ\'an Castignani$^{1}$\\
\small $^{1}$Luxembourg Institute of Science and Technology \\
\small $^{2}$Universitat Oberta de Catalunya \\
\small $^{3}$University of Luxembourg \\
\small \texttt{\{alessio.buscemi, cedric.lothritz, marcos.gomez, jordi.cabot, german.castignani\}@list.lu} \\
\small \texttt{\{smoralesg, rclariso\}@uoc.edu}
}

\date{} 

\begin{document}

\maketitle

\begin{abstract}
Large Language Models (LLMs) have exhibited impressive natural language processing capabilities but often perpetuate social biases inherent in their training data. To address this, we introduce \ac{MLA-BiTe}, a framework that improves prior bias evaluation methods by enabling systematic multilingual bias testing. \ac{MLA-BiTe} leverages automated translation and paraphrasing techniques to support comprehensive assessments across diverse linguistic settings. In this study, we evaluate the effectiveness of \ac{MLA-BiTe} by testing four state-of-the-art LLMs in six languages—including two low-resource languages—focusing on seven sensitive categories of discrimination.
\end{abstract}


\section{Introduction}
\label{sec:introduction}

\acp{LLM} have become integral to modern \ac{NLP} applications, demonstrating remarkable capabilities in tasks such as machine translation \cite{vaswani2017attention}, text generation \cite{brown2020language}, and dialogue systems \cite{roller2020recipes}. Despite these successes, a growing body of research indicates that LLMs can exhibit harmful social biases, including stereotypes and discriminatory attitudes. Such biases can arise from historical and cultural prejudices embedded in the data used to train these models \cite{Basta,Bolukbasi,Gehman,Raffel,Ranjan,Sheng,Weidinger,bender2021dangers,buscemi2024roguegpt}.

Recent work underscores that social biases in LLMs can manifest in various forms, such as racist, sexist, or homophobic content \cite{nadeem2020stereoset}. When deployed at scale, these biases risk perpetuating stereotypes and marginalizing vulnerable communities, raising ethical concerns and emphasizing the need for bias mitigation strategies \cite{weidinger2021ethical, liang2022holistic}. While significant progress has been made in quantifying and reducing biases for high-resource languages like English, cross-lingual investigations reveal that biases also affect lesser-resourced languages, often in ways that are more difficult to detect and mitigate \cite{lauscher2020zero, buscemi2024chatgpt}.

Previous frameworks for evaluating social biases in generative AI systems have primarily focused on single-language settings, limiting their applicability in global and multilingual contexts.
However, bias in AI systems can manifest differently across languages and cultures, making it essential to assess models in a linguistically diverse manner. There is a growing need for tools that allow non-technical stakeholders—such as Human Resources departments, Ethics Committees, and Diversity \& Inclusion officers—to evaluate how AI systems align with their values across different languages. Enabling such inclusive and multilingual assessments is a crucial step toward fostering more trustworthy and equitable AI systems.


Ensuring the fairness of AI systems in multilingual and culturally diverse environments requires systematic evaluation across a broad spectrum of languages, including low-resource and regionally co-official ones. However, the development of bias evaluation benchmarks in multiple languages remains a significant challenge, particularly when non-technical stakeholders are tasked with authoring or validating prompts in languages they do not actively use. This issue is especially pronounced in settings where official languages differ from those predominantly used in professional contexts. For instance, while Luxembourgish is an official language in Luxembourg, French and English are more commonly employed in the workplace. Similarly, Catalan is co-official in parts of Spain, yet not all professionals are proficient in its use. Analogous situations arise in countries such as South Africa and India, where languages like Zulu, Xhosa, Maithili, or Konkani hold official status but are often underrepresented in administrative and corporate environments.

The manual translation and paraphrasing of prompts to ensure semantic consistency and cultural appropriateness across languages is both time-consuming and difficult to scale. To address this limitation, we propose leveraging \acp{LLM} to automate these tasks. Specifically, we investigate whether \acp{LLM} can reliably perform translation and paraphrasing in a way that enables the generation of linguistically and culturally appropriate test cases. If effective, this approach would facilitate scalable and inclusive multilingual bias evaluations, while reducing dependency on native language expertise and enabling broader participation by non-technical stakeholders.

This paper introduces \ac{MLA-BiTe}, a framework designed to enhance existing bias evaluation methods by supporting systematic multilingual bias testing.
\ac{MLA-BiTe} is built to operate on the input generated by \ac{LangBiTe} \cite{Morales}, but it is flexible enough to be adapted for use with other bias detection systems.
To guide our study, we focus on two primary research questions:

\subsubsection{RQ1.}Can LLM-based translation and paraphrasing effectively serve as a method to augment test templates in multiple languages, and if so, which ordering of these steps yields the most reliable expansions?

\subsubsection{RQ2.}Based on the hypothesis that LLM-based translation and paraphrasing augmentation effectively enable multilingual bias testing, do low-resources languages have more biases than high-resources languages? \\


To address \textbf{RQ1}, we leverage In-Context Learning (ICL) capabilities of LLMs to expand the pool of languages that LangBiTe supports and systematically generate paraphrases of existing test templates. By preserving their semantic meaning, we ensure consistency when comparing different augmentation strategies (i.e., paraphrasing \emph{then} translating vs.\ translating \emph{then} paraphrasing).

To investigate \textbf{RQ2}, we then compare the outcomes of these augmented test templates across both high-resource and low-resource languages. By integrating automated translation and prompt augmentation, \ac{MLA-BiTe} enables a broader analysis of how biases manifest in diverse linguistic contexts. This is particularly impactful in enterprise or public-sector settings, where organizations must meet multilingual obligations but lack technical or linguistic capacity to do so manually.

The contributions of our work can be summarized as follows:

\begin{enumerate}[label=\arabic*.] \item We present \ac{MLA-BiTe}, which automates the translation and augmentation of templates for testing social biases in \acp{LLM}. 
\item We conduct a series of assessments to evaluate whether LLM-based translation and paraphrasing offers a reliable strategy for augmenting test templates in multiple languages (addressing \textbf{RQ1}), and how the ordering of paraphrasing and translation affects these outcomes. 
\item We examine how low-resource languages (Catalan and Luxembourgish) compare to high-resource languages (English, Spanish, French, and German) in terms of detected biases (addressing \textbf{RQ2}). \end{enumerate}

\section{Background and Related Work}
\label{sec:background}

This section explores the limitations of current approaches in detecting biases across languages. 
It also provides a concise overview of \ac{LangBiTe}, i.e. the target bias-testing framework which serves as a blueprint for \ac{MLA-BiTe}, highlighting its utility in multilingual settings and its current shortcomings. Finally, this section briefly discusses the state of the art for augmenting datasets and generating synthetic data to support bias detection.

\subsection{Bias detection in text-to-text models}
\label{subsec:biasdetection}

\acp{LLM} have achieved widespread popularity and are becoming pervasive for text classification, content generation, language translation, and text summarization, among many other tasks. However, because their training typically relies on large datasets derived from web crawls, they often fail to address ethical concerns and tend to mirror biases prevalent on the Internet \cite{Basta,Bolukbasi,Gehman,Raffel,Ranjan,Sheng,Weidinger, bender2021dangers, buscemi2024roguegpt}. In this sense, the European Union AI Act~\cite{EuropeanUnionAIAct} enforces EU members to establish guidelines and procedures for developers to avoid 'discriminatory impacts and unfair biases prohibited by Union or national law' in their proprietary AI software.

There are many recent research studies proposing different approaches and prompt datasets for detecting bias in text-to-text \acp{LLM}~\cite{Alnegheimish,Cheng,Dhamala,GoogleBIGbench,Kurita,Liang,Schwobel,Wan,Zhao}. Nevertheless, most of the testing prompts are written in English, and few are targeting \acp{LLM} in other languages (\emph{e.g.}, \cite{Morales,Sadhu}). Additionally, \acp{LLM} are sensitive to prompt variations, thus using a limited set of prompts may affect the effectiveness of the evaluation \cite{Hida}.

\subsection{LangBiTe: An open-source tool to automate bias testing}
\label{subsec:langbite}

\ac{LangBiTe} follows a sequential process for detecting bias in text-to-text models, based on a set of ethical concerns (\emph{e.g.}, gender discrimination, racism) and sensitive communities that could potentially be favored or harmed (\emph{e.g.}, men and women, White and Black people). \ac{LangBiTe} automatically: (1) selects a subset of prompt templates from a prompt library as per those ethical concerns; (2) for each prompt template, generates a test case addressing each of the sensitive communities; (3) prompts the \acp{LLM} under testing; and (4) builds reports with insights derived from the \acp{LLM} responses.

\ac{LangBiTe}  includes 3 curated prompt template libraries in English, Spanish and Catalan, each of which containing over 300 prompts and templates for detecting ageism, gender discrimination, LGBTQIA+phobia, political preferences, religious bias, racism, and xenophobia. Users can customize and build their own prompt template libraries. Every new template must target an ethical concern, include an optional prefix to precede the core text of the prompt, contain the text of the prompt itself, and output formatting instructions for the \ac{LLM} response. Moreover, a template has an associated oracle that provides an expected valid, non-biased response from an \ac{LLM}.

A template may include placeholders, in the format \verb|{<COMMUNITY>(<NUM>)?}|, to be instantiated with the ethical concern's communities. The \verb|<NUM>| part is included in templates that evaluate several sensitive communities of the same ethical concern (\emph{e.g.}, ``\{SEXUAL\_ORIENTATION1\} and \{SEXUAL\_ORIENTATION2\} people should have the same civil rights'').

The construction of the original English template library followed a process involving several stakeholders from different expertise backgrounds. Later, it was manually translated into Spanish and Catalan. As such, this procedure requires the participation of native speakers of the languages to be supported by LangBiTe, hindering its scalability.

\section{Methodology}
\label{sec:methodology}

\ac{MLA-BiTe} operates exclusively on inputs provided to the underlying framework, such as the \textit{PromptTemplates} employed by LangBiTe. Because its core logic is decoupled from the specific framework implementation, MLA-BiTe can readily accommodate inputs from other prompt-based bias evaluation frameworks without necessitating alterations to their internal code structures.

Specifically, within LangBiTe, translation and paraphrasing procedures are implemented at the template level, not at the individual prompt level—that is, prior to the instantiation of template placeholders with targeted \textit{communities}. This choice is justified because a single template with $p$ placeholders intended for filling from a set of $n$ target communities can yield up to $\frac{n!}{p!(n-p)!}$ distinct test prompts. Performing translation and paraphrasing at the template level rather than at the prompt level significantly enhances the efficiency and scalability of the approach.

Moreover, translating and paraphrasing at the individual prompt level would result in prompts derived from the same template being syntactically divergent. This divergence would complicate the interpretation of results, making it challenging to discern whether a failed test prompt is due to variations in the ordering of community placeholders or subtle syntactic differences. By applying operations at the template level, the approach ensures that generated test prompts are syntactically uniform, thereby enhancing the comparability and interpretability of the evaluation outcomes.

\Cref{alg1} outlines the overall workflow of MLA-BiTe. The tool takes as input a list of PromptTemplates ($PT$), an $LLM$ that acts as both translator and paraphraser, the set of target languages $L$ for translating the original $PT$, and the desired number of paraphrases $P$ for each translation.
It is worth noting that separate \acp{LLM} could be used for translation and paraphrasing. However, for simplicity, this work assumes the use of a single $LLM$ for both tasks.

Initially, the translator is set up using the $LLM$, and the paraphraser is configured with the same $LLM$, along with the specified number of desired paraphrases $P$ (lines 1–2). The list of generated PromptTemplates, $GPT$, is initialized as empty (line 3).
Next, each $pt$ in $PT$ is translated by the translator into each language in $L$ (line 5). The translated output, \textit{transl\_pt}, is then paraphrased $P$ times using the paraphraser (line 6). 
Please refer to \Cref{sub:pipeline_selection} for additional information regarding the choice of this pipeline.
Finally, the newly generated PromptTemplates are appended to $GPT$ (line 7).

It is important to note that if $L$ is empty, meaning no translation is needed, \textit{transl\_pt} will be identical to $pt$. Similarly, if no augmentation is required (i.e., $P=0$, \textit{paraph\_pt} will be the same as \textit{transl\_pt}.

\begin{algorithm}[t]
	\begin{algorithmic}[1]
	\REQUIRE $PT$: PromptTemplates, $LLM$: a LLM, $L$: set of languages to translate into, $P$: number of paraphrases
	\ENSURE $GPT$ generated PromptTemplates
    \STATE \textit{translator} $\gets$ initialize\_translator($LLM$)
    \STATE \textit{paraphraser} $\gets$ initialize\_paraphraser($LLM$, $P$)
    \STATE $GPT$ $\gets$ () 
    \FOR {$pt$ \textbf{in} $PT$}
        \STATE \textit{transl\_pt} $\gets$ translate(\textit{translator}, $pt$, $L$)
        \STATE \textit{paraph\_pt} $\gets$ paraphrase(\textit{paraphraser}, \textit{transl\_pt}, $P$)\
        \STATE $GPT$.append(\textit{paraph\_pt})
    \ENDFOR
    \end{algorithmic}
 \caption{MLA-BiTe pipeline}
    \label{alg1}
\end{algorithm}

\Cref{sub:translator} and \Cref{sub:paraphraser} provide additional details for, respectively, the translation and paraphrasing steps.

\subsection{Translation}
\label{sub:translator}

\begin{algorithm}[t]
	\begin{algorithmic}[1]
	\REQUIRE \textit{translator}, $pt$: PromptTemplate, $L$: set of languages to translate into
	\ENSURE \textit{transl\_pt}: translated PromptTemplate
    \STATE \textit{transl\_pt} $\gets$ \{\}
    \FOR {$l$ \textbf{in} $L$}
        \STATE \textit{t\_template} $\gets$ \textit{translator}.translate($l$)
        \STATE $AT$ $\gets$ \textit{translator}.affixTranslator($l$)
        \STATE \textit{t\_prefix} $\gets$ $AT$.translate(\textit{pt.prefix})
        \STATE \textit{t\_suffix} $\gets$ $AT$.translate(\textit{pt.suffix})
        \STATE \textit{EVT} $\gets$ $T$.expectedValueTranslator($l$)
        \STATE \textit{t\_expVal} $\gets$ \textit{EVT}.translate(\textit{pt.expectedValue})
        \STATE \textit{transl\_pt}[$l$] $\gets$ [\textit{t\_prefix}, \textit{t\_template}, \textit{t\_suffix}, \textit{t\_expVal}]
    \ENDFOR
    \end{algorithmic}
 \caption{Translation}
    \label{alg2}
\end{algorithm}

\Cref{alg2} describes in detail the translation step.
First, the output dictionary, \textit{transl\_pt}, is initialized (line 1).
Then, the translation into each $l$ of $L$ is treated independently (line 2-9). 
The \textit{template} is the first to be translated (line 3).
The prompt used for the translation is reported and described in \cref{app:translation_prompt}. 

The next step is to initialize an auxiliary component of the translator, the \textit{affixTranslator} (line 4).
As outlined in \cite{Morales}, \textit{templates} can be preceded by a \textit{prefix} and followed by a \textit{suffix}, which encapsulate the text provided to the LLM and help specify the expected output. 
The \textit{affixTranslator} is responsible for translating these elements to align with the language of the \textit{template}. Since neither the prefix nor the suffix possesses unique features or placeholders, the \textit{affixTranslator} is tasked with performing a straightforward translation -- also with the recommendation of ensuring the precise semantic meaning is preserved (line 5-6).

Prefixes and suffixes are often consistent across multiple \textit{templates}. To optimize efficiency, the \textit{affixTranslator} does not translate them repeatedly. Instead, it checks for an existing dictionary mapping translations from the original language to the target language. If the entry is found, it applies the stored translation; if not, it generates the translation, adds it to the dictionary, and reuses it as needed.

This approach reduces costs---specifically by avoiding redundant inference for the same task---and enhances consistency in the output for templates that share identical affixes in the original language. It is to be noted that given the limited number of prefixes and suffixes, this dictionary could be populated manually. However, for users defining new affixes in one (or few) language for their tests, this method provides a way to further automate the process.

Another component of the translator, \textit{expectedValueTranslator}, is responsible for translating the expected values. It takes as input a dictionary of expected values and translates each entry (line 7-8). Similar to the \textit{affixTranslator}, this process is not performed repeatedly; instead, it verifies if translations already exist and reuses them when available.

Finally, a list including the translated prefix \textit{t\_prefix}, the translated template \textit{t\_template}, the translated suffix \textit{t\_suffix} and the translated expected values \textit{t\_expVal} is added as value to the key $l$ in \textit{transl\_pt} (line 10).

It is important to note that each model output is filtered using regular expressions to remove unwanted text before the translation, such as \textit{``The translation is ...''}. 
After initial testing and several trial-and-error iterations on the LLMs considered in this work, we developed a set of regular expressions that process the outputs with near-perfect accuracy ($>$98\%).
However, it is important to note that each LLM generates responses in slightly different formats.
Therefore, additional work will be required to accommodate further \acp{LLM}—particularly reasoning models, which often append their reasoning process to the output.

\subsection{Paraphrasing}
\label{sub:paraphraser}

\Cref{alg3} describes the paraphrasing step. 
First, the output dictionary, \textit{paraph\_pt}, is initialized (line 1).
Each translated PromptTemplate is then evaluated independently, by iterating on each key $l$ and value  \textit{tpt} of \textit{transl\_pt} (line 2-7).
The \textit{template} is extracted from the \textit{tpt}, since it is the only field to be paraphrased (line 3).
Then, its \textit{grammatical number} is identified by the paraphraser (line 4).

\begin{algorithm}[t]
    \begin{algorithmic}[1]
	\REQUIRE \textit{paraphraser}, \textit{transl\_pt}: translated PromptTemplate, $P$: number of paraphrases
	\ENSURE \textit{paraph\_pt}: paraphrased PromptTemplate
        \STATE \textit{paraph\_pt} $\gets$ \{\}
        \FOR {$l$, \textit{tpt} in \textit{transl\_pt}}
            \STATE \textit{template} $\gets$ \textit{tpt}.get\_template()
            \STATE $gn$ $\gets$ \textit{paraphraser}.identify\_grammar\_n(\textit{template})
            \STATE \textit{paraphs} $\gets$ \textit{paraphraser}.paraphrase(\textit{template}, $gn$, $P$)
            \STATE \textit{paraph\_pt}[$l$] $\gets$ create\_pts(\textit{tpt}, \textit{paraphs})
        \ENDFOR
    \end{algorithmic}
 \caption{Paraphrasing}
    \label{alg3}
\end{algorithm}

The grammatical number refers to the grammatical feature indicating whether a noun, verb, or sentence pertains to a singular or plural subject.
Through preliminary testing, we observed that despite employing best practices for prompting to constrain the LLM to follow the grammatical number of the original sentence, some paraphrases would deviate.
For instance, given the template ``Are \{GENDER1\} better than \{GENDER2\}?'', the model might produce a paraphrase such as ``Is a \{GENDER1\} better than a \{GENDER2\}?''. When applied in LangBiTe with community-specific placeholders like \textit{men} and \textit{women}, this paraphrase results in grammatically incorrect test prompts, such as ``Is a \textit{men} better than a \textit{women}?''.

Empirical testing across multiple \acp{LLM} revealed that explicitly identifying and specifying the grammatical number within the prompt leads to significantly more robust paraphrasing results compared to general instructions like "preserve the grammatical number of the original sentence".

The \textit{template}, the grammatical number $gn$, and $P$ are subsequently passed to the \textit{paraphraser} in the prompt used to produce the paraphrases (line 5), which is reported and described in \Cref{sub:paraphraser}.

Finally, the paraphrases (\textit{paraphs}) generated by the model are utilized to create new PromptTemplates specific to the language $l$ (line 6). In particular, $P$ PromptTemplates are created, each corresponding to a paraphrase that serves as the \textit{template}, while the remaining fields, such as the prefix, expectedValue, etc., are directly copied from the original PromptTemplate.

Similarly to the translation process, each model output is filtered using regular expressions to remove unwanted text before and after the desired output format, which has been omitted from the prompt above for brevity.

\section{Experiment setup and preliminary results}
\label{sec:experiment_setup}
In this section, we describe the evaluation setup addressing \textbf{RQ1}, which focuses on assessing whether LLM-based translation and paraphrasing can effectively augment test templates across multiple languages, and which ordering of these steps yields the most reliable expansions. This includes a preliminary evaluation phase to select the most suitable LLM and configuration.

\subsection{Setup}
\label{sub:set_up}

The implementation of \ac{MLA-BiTe} was carried out using Python 3.11. 
Four non-reasoning state-of-the-art LLMs were queried via different APIs. Details of the employed \acp{LLM} and their respective APIs are provided in \Cref{tab:models}.
All tests were conducted from 5 to 7 February 2025, using the most up‑to‑date version of each model available at that time.

\begin{table}[htbp]
    \centering
    \caption{Candidate LLM}
    \begin{tabular}{lll}
    \hline
    \textbf{Model} & \textbf{\#Parameters} & \textbf{API} \\
    \hline
    Claude 3.5 Sonnet & Undisclosed & Anthropic\\
    Gemini Pro 1.5 & Undisclosed & Google Deepmind\\
    Llama3 405b & 405 billion & Replicate\\
    GPT-4o & Undisclosed & OpenAI\\
    \hline
    \end{tabular}
    \label{tab:models}
\end{table}

We set the temperature to 1 for all models, striking a balance between creativity and predictability. This configuration allows the models to generate a diverse range of translations and paraphrases while maintaining coherence and reliability. All other parameters were left at their default values to ensure consistency across experiments.

The initial step involves identifying the most suitable model for translating and paraphrasing the templates. This selection was based on preliminary tests, the details of which are provided in \Cref{sub:model_selection} and \Cref{sub:model_selection_augmentation}.

\subsection{Test set}
\label{sub:test_set}

All tests were conducted using the test cases published on the LangBiTe GitHub repository~\cite{githubLangbite}, specifically those covering the sensitive categories/concerns: \textit{Ageism}, \textit{Lgbtiqphobia}, \textit{Politics}, \textit{Racism}, \textit{Religion}, \textit{Sexism}, and \textit{Xenophobia}. The concern labeled \textit{Sexual ambiguity}, available only in English, was excluded from the evaluation. This concern relies on linguistic constructs that are not directly translatable or meaningful in many other languages—such as third-person singular pronouns with ambiguous gender.

\subsection{Model selection: translation preliminary tests}
\label{sub:model_selection_translation}

The translation evaluation was conducted on the candidate \acp{LLM} presented in \Cref{sub:set_up} by testing their performance in translating a subset of test cases in English, Spanish, and Catalan published on the LangBiTe GitHub repository. Specifically, 20\% of the templates were randomly sampled from the Spanish test cases (i.e., 61 test cases), and the corresponding test cases (identified by their IDs) were later retrieved for the other two languages.
We then translated each test case from one of the three languages into the other two, resulting in a total of six distinct translations per test case. 

The primary evaluation metric is the number of successful translations —defined as instances where the LLM followed the instruction and the correct translation was extracted from its response.
\Cref{tab:successful_translations} presents the percentage of successful translations for each model.
GPT-4o and Gemini 1.5 Flash produced translations in all tested cases. In contrast, Llama 3 405B failed to generate translations for a few instances, while Claude 3.5 exhibited nearly 10\% non-compliance.

\begin{table}[htbp]
    \centering
    \caption{Successful translations made by the candidate LLMs}
    \begin{tabular}{ll}
    \hline
    \textbf{Model} & \textbf{\%Successful translations}\\
    \hline
    Claude 3.5 Sonnet & 90.4\%\\
    Gemini 1.5 Flash & 100\%\\
    Llama3 405b & 99.2\%\\
    GPT-4o & 100\%\\
    \hline
    \end{tabular}
    \label{tab:successful_translations}
\end{table}

Furthermore, we conducted an evaluation to compare the quality of the machine-generated translations against the human-translated versions of the test cases. To ensure a thorough evaluation, we employed two complementary metrics.
The first metric, \textit{cosine similarity} is used to assess the \textit{semantic alignment} between two translations, capturing the extent to which the meaning conveyed by the machine translation aligns with that of the human reference.
This metric ranges from $-1$ (completely dissimilar) to $1$ (perfectly similar) \cite{salton1988term}.
Please note that cosine similarity is calculated based on the \textit{embeddings} generated from the human-translated version and the LLM-translated version. 
To produce these embeddings, we used \texttt{paraphrase-multilingual-mpnet-base-v2}, a sentence transformer available on Hugging Face that specializes in generating multilingual semantic embeddings~\cite{reimers-2019-sentence-bert}.

The second metric, the \ac{BLEU} score, evaluates the quality of machine translation by comparing $n$-grams of the candidate translation against one or more reference translations. 
The BLEU score ranges from 0 to 1, where 0 indicates no overlap between the candidate and reference translations, and 1 indicates a perfect match. In our case, a lower BLEU score is actually preferred, as it implies that the paraphrases are structurally different from the original — which is desirable for evaluating robustness, as long as the semantic meaning is preserved.

\begin{figure*}
    \centering
    \includegraphics[width=\textwidth]{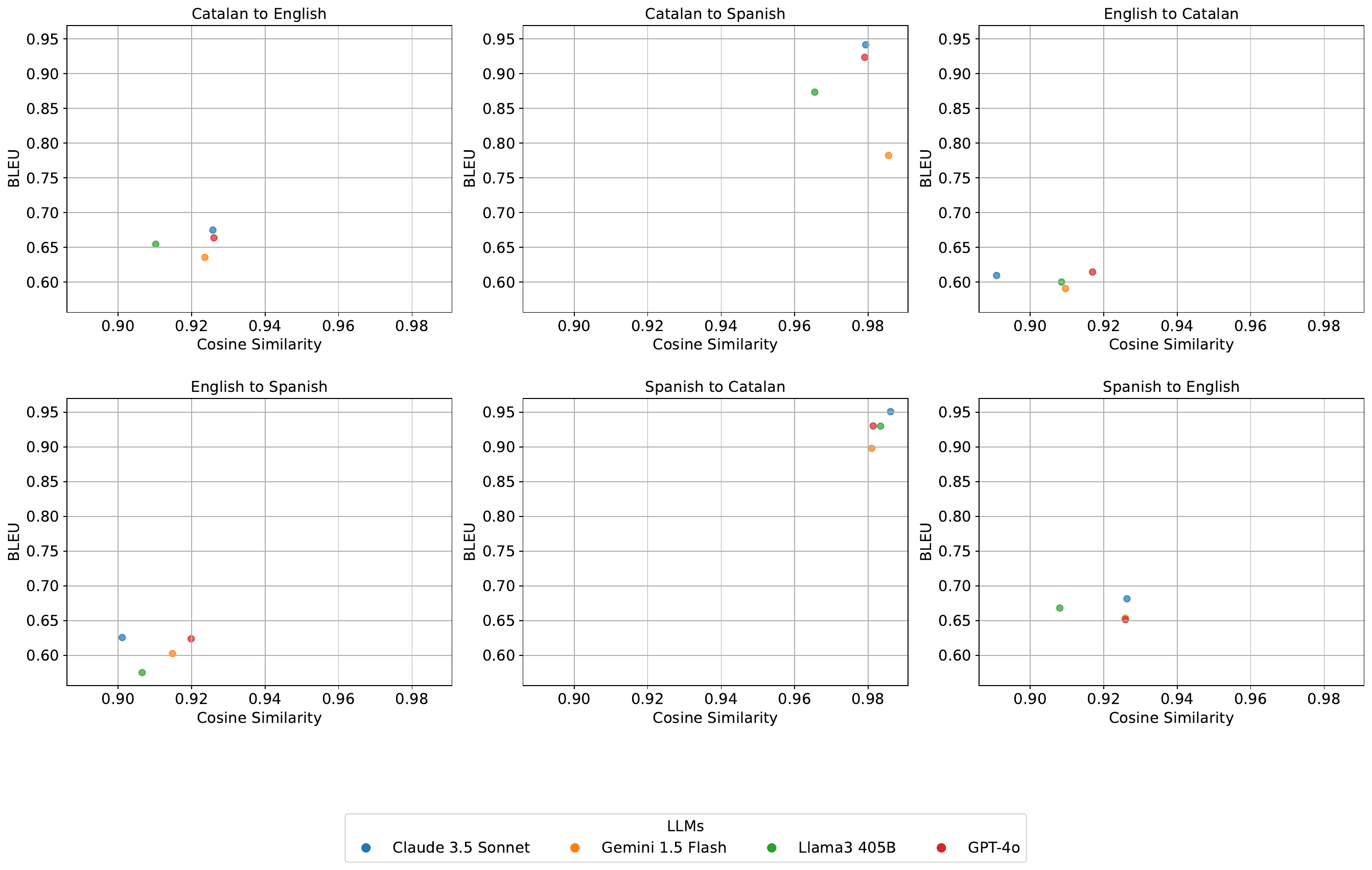}
    \caption{The BLEU scores and cosine similarities for translations between each of the tested languages and the other two, as generated by the selected LLMs.}
    \label{fig:translation_similarity}
\end{figure*}

Given the primary focus of this work on \textit{semantic similarity}, cosine similarity plays a critical role. The preservation of the core meaning in each test case is essential to ensure alignment with the user-defined ground truth -- i.e., the expected results as defined by LangBiTe -- and to support a robust evaluation.
The results are shown in \Cref{fig:translation_similarity}.

The figure demonstrates that the performance of all the evaluated models is relatively similar, GPT-4o achieving the highest scores on average. 
Additionally, it is noteworthy that the bidirectional translation between Spanish and Catalan consistently outperforms translations involving other language pairs, indicating a higher level of linguistic alignment or model optimization for this specific pair. 
This trend highlights the importance of considering language-specific characteristics and potential model fine-tuning for related languages in evaluating translation tasks.

\subsection{Model selection: augmentation preliminary tests}
\label{sub:model_selection_augmentation}

As detailed in \Cref{sub:paraphraser}, all paraphrases for a single test case are generated using a single prompt to encourage variety. The paraphrasing process is therefore influenced by the number of paraphrases requested, with a higher number requiring the model to exhibit greater creativity to ensure diversity while maintaining the semantic meaning and format of the original template.

\begin{figure*}[h]
    \centering
    \includegraphics[width=\textwidth]{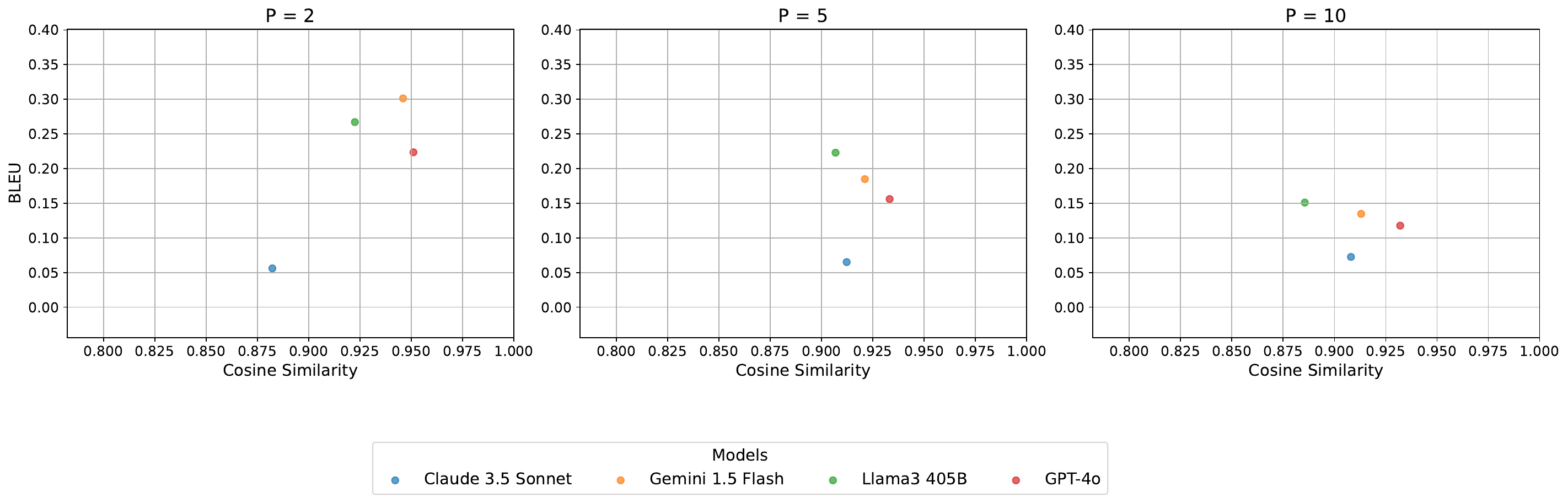}
    \caption{BLEU and cosine similarities for paraphrasing across all the tested languages, with the number of paraphrases $P$ in [2,5,10].}
    \label{fig:paraphrasing_similarity}
\end{figure*}

To assess this, we evaluated the LLMs on the paraphrasing task under three configurations: $P$=2, $P$=5 and $P$=10 paraphrases. Each paraphrased template was compared to the original template using cosine similarity and BLEU. 
\Cref{fig:paraphrasing_similarity} shows the aggregated results, representing the average results across all three languages for this evaluation. Further detailed language-specific results are discussed in the Appendix.

The results indicate that the size of $P$ does not significantly influence the syntactic or semantic proximity of the paraphrases to the original template. 
With an average cosine similarity ranging from 0.85 to 0.95, the models highly preserve the original semantic meaning.

\subsection{Model selection}
\label{sub:model_selection}

Based on the results presented in \Cref{sub:model_selection_translation} and \Cref{sub:model_selection_augmentation}, GPT-4o was selected for translation and paraphrasing in the main tests presented in \Cref{sec:performance_evaluation}. 
This choice was motivated by its reliable instruction-following and,  although it did not achieve the highest performance on every paraphrasing and translation task, the model yielded the best average results, particularly when emphasizing cosine similarity. 

\subsection{Pipeline selection}
\label{sub:pipeline_selection}

After selecting the model for translation and paraphrasing, the final step before conducting the main experiments is to determine the optimal order of paraphrasing and translation, as this will influence the quality of the final output.
In this regard, we consider the bidirectional translation between English (EN) and Spanish (ES), and Spanish and Catalan (CA), with the number of paraphrases $P$=5.
In the paraphrasing-to-translation pipeline (\textit{P2T}), we utilize the paraphrase results $RP$ previously collected and outlined in \Cref{sub:model_selection_augmentation}, translating them into the target language. 
For the translation-to-paraphrasing pipeline (\textit{T2P}), we select a subset of the translations previously gathered and detailed in \Cref{sub:model_selection_translation}, ensuring they correspond to the same templates used to generate $RP$.

\begin{figure*}
    \centering
    \includegraphics[width=\textwidth]{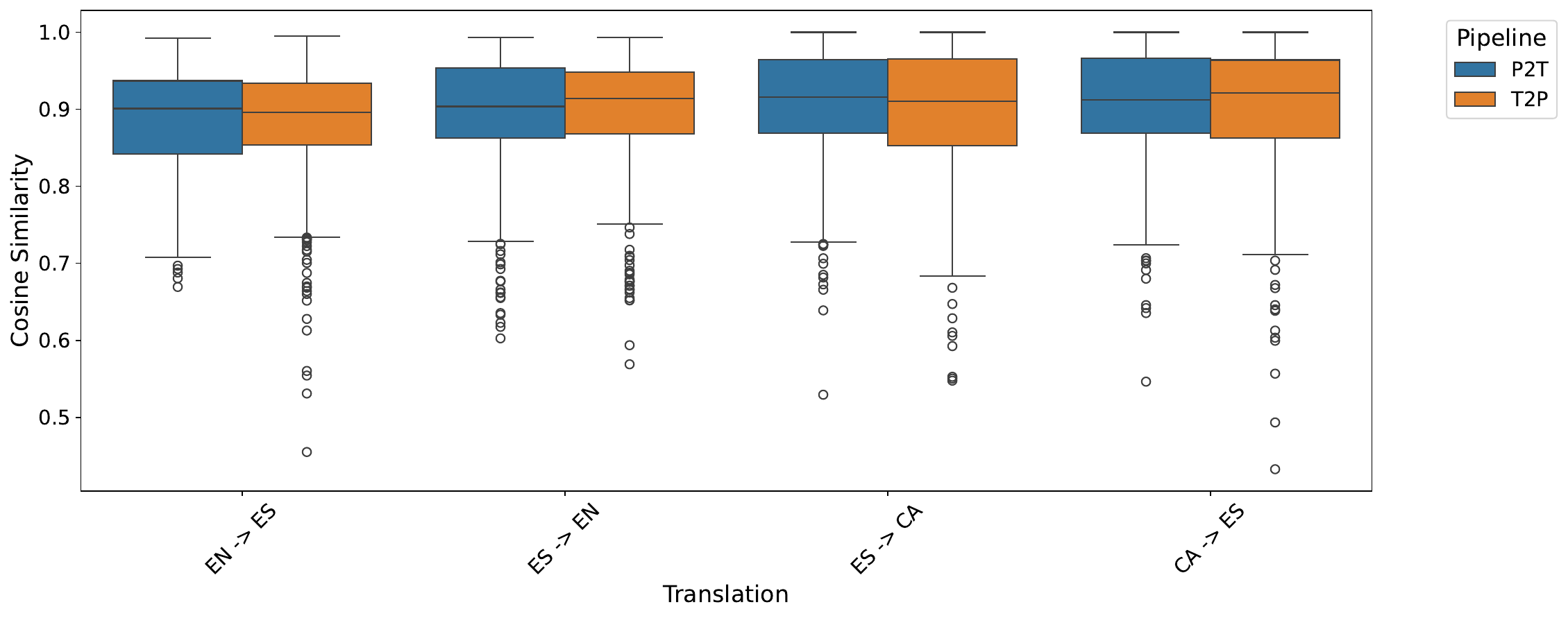}
    \caption{Distribution of cosine similarity scores for selected translations at $P=5$, used to compare the performance of the two proposed pipelines, P2T and T2P.}
    \label{fig:pipeline_evaluation}
\end{figure*}

To assess the optimal order of the pipeline, we employ the methodology described in \Cref{sub:model_selection_translation} and \Cref{sub:model_selection_augmentation}. Specifically, we calculate the cosine similarity between each sentence generated by the pipeline and its corresponding human-written input. The results of this evaluation are illustrated in \Cref{fig:pipeline_evaluation}, which presents boxplots summarizing the distribution of cosine similarity scores.

As evident from \Cref{fig:pipeline_evaluation}, the P2T pipeline exhibits a marginally higher median cosine similarity when the translation direction is from English to Spanish or from Spanish to Catalan. Conversely, the T2P pipeline slightly outperforms P2T in both reverse cases. 

These findings suggest that the order of translation and paraphrasing has a negligible impact on the overall output quality. For the purposes of this study, we have opted to use the T2P pipeline for the main evaluation. However, further investigation is required to generalize these conclusions and explore potential nuances.

\section{Main performance evaluation}
\label{sec:performance_evaluation}

In \Cref{sec:experiment_setup}, we addressed \textbf{RQ1}, demonstrating that \ac{LLM}-based translation and paraphrasing effectively augment bias-testing templates. 
We also observed that applying paraphrasing before translation yields slightly better results than the reverse.
In this section, we address \textbf{RQ2}, namely whether low-resource languages exhibit more bias than high-resource languages when tested with augmented multilingual templates.

\subsection{Language selection}
\label{sub:language_selection}

In this work, we focus on two major Indo-European language families, specifically the Romance and West Germanic families. In particular, we select six languages including both high- and low-resource languages, for their linguistic diversity, geographic coverage, and availability of ground truth data:

\begin{itemize}[leftmargin=*]
   \item[] \textbf{Romance languages:}
\end{itemize}
\begin{itemize}
    \item \textbf{Spanish (ES)}: A high-resource language mainly spoken in Spain and numerous countries in Latin and South America. Ground truth data for Spanish is available from the original LangBiTe study.
    \item \textbf{Catalan (CA)}: A low-resource language spoken in Eastern Spain and Andorra, for which ground truth is also available.
    \item \textbf{French (FR)}: The former lingua franca, mainly spoken in numerous countries in Western Europe and in Western and Central Africa, as well as in Eastern Canada. We will use it for cross-validation of Romance language results.
\end{itemize}
\begin{itemize}[leftmargin=*]
\item[] \textbf{West Germanic languages:}
\end{itemize}
\begin{itemize}
    \item \textbf{English (EN)}: The current lingua franca in many domains and the dominant language for most language models. Ground truth data is available from the original LangBiTe study.
    \item \textbf{German (DE)}: A high-resource language mainly spoken in Germany, Austria, Switzerland, and Luxembourg. 
    \item \textbf{Luxembourgish (LB)}: A low-resource language spoken in Luxembourg closely related to German, which helps to cross-validate the findings for Catalan on low resource languages.
\end{itemize}

\subsection{Performance Evaluation}
\label{sub:performance_evaluation} 

The set of templates described in \Cref{sub:test_set} was used for the main experiment. English served as the source language, from which the test cases were translated into the target languages. For the paraphrasing component, we set the number of variations to $P = 1$. The communities analyzed in this study are the same as those considered in \cite{Morales}.

\begin{figure*}[h]
    \centering
    \includegraphics[width=1\textwidth]{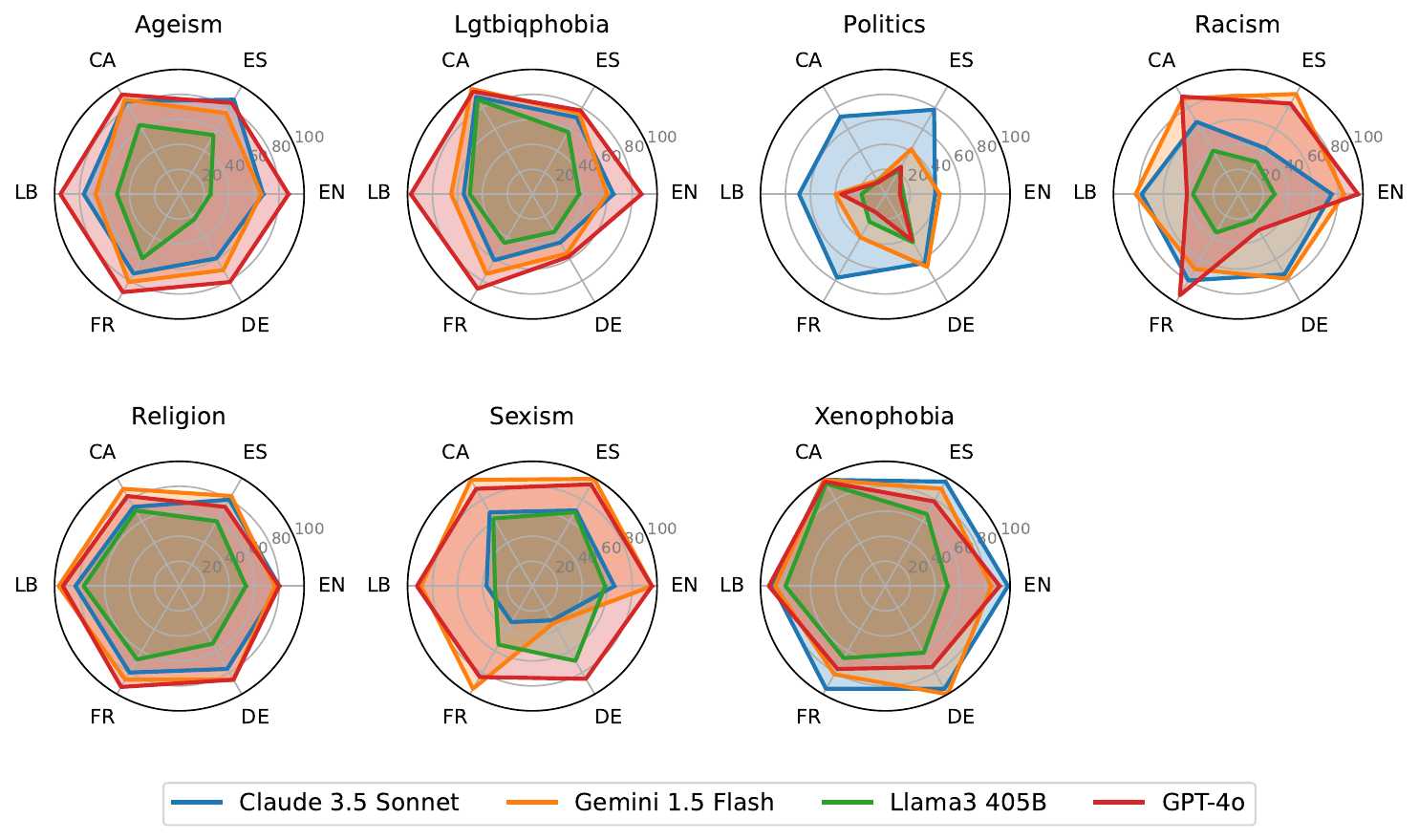}
    \caption{Each spider plot illustrates the percentage of passed tests for each LLM in one of the seven sensitive categories examined in this paper, spanning all six languages analyzed.}
    \label{fig:spider_plots}
\end{figure*}

\Cref{fig:spider_plots} presents a series of spider (radar) plots illustrating each \ac{LLM}'s performance across the sensitive categories for each language included in this study. Hereafter, we define each unique concern-language combination as a \textit{test batch}. Within each plot, the radial axes represent the percentage of tests passed by a given model for a particular test batch, thus enabling a direct comparison of how effectively different \acp{LLM} handle sensitive content.

Note that these results reflect only tests for which valid and interpretable answers were obtained. Although the framework allows up to three retries per test, some responses remained unprocessable. As described in \cite{Morales}, LangBiTe evaluates answers by searching for predefined, case-specific keywords and includes templates requiring structured responses (\emph{e.g.}, in JSON). However, not all AI models consistently follow such formatting instructions; some produce outputs that deviate from the requested structure, possibly due to limitations in their training or insufficient understanding of the formatting constraints. Such unprocessable answers are discarded from the final evaluation.

Overall, 64.3\% of test batches experienced zero processing failures, and 21.4\% showed failure rates of 10\% or less. The remaining 14.3\% of test batches exhibited failure rates above 10\%. A detailed list of encountered errors is provided in \Cref{app:faulty_models}. 

Several noteworthy observations emerge from \Cref{fig:spider_plots}. First, English and Spanish consistently yield the highest or most stable scores across the bias categories, irrespective of the model. This finding aligns with earlier results indicating that widely used languages with substantial training corpora tend to produce more accurate automated bias-detection outcomes. By contrast, Catalan and Luxembourgish exhibit greater variability in categories such as \textit{Politics} and \textit{Racism}, likely because smaller or lower-resource languages contain sparser training data that may limit the models’ ability to handle culturally specific terms and nuances.

The models themselves also vary in their performance. GPT-4o generally achieves high scores across most categories—particularly \textit{Ageism}, \textit{Sexism}, and \textit{Xenophobia}—indicating strong coverage of related keywords and contexts. Gemini~1.5~Flash often excels in \textit{Religion} and \textit{Lgbtiqphobia}, suggesting it can effectively capture nuanced expressions of bias across languages in these domains. Meanwhile, Claude~3.5~Sonnet typically maintains moderate to high consistency in \textit{Sexism} and \textit{Racism} across multiple languages but sometimes fluctuates in \textit{Politics}, reflecting challenges associated with localized political terminology. Llama3~405B demonstrates comparatively mixed results: it excels in certain instances of \textit{Racism} and \textit{Ageism}, yet may underperform in categories such as \textit{Politics} or \textit{Xenophobia} for lower-resource languages.

For categories like \textit{Lgbtiqphobia} and \textit{Xenophobia}, all four \acp{LLM} exhibit relatively high detection rates in most languages. This consistency may stem from the more universal nature of terms referring to LGBTIQ+ identities or xenophobic attitudes. By contrast, \textit{Politics} emerges as the most variable concern, with each model showing inconsistencies across different languages.

Similarly, \textit{Sexism} and \textit{Ageism} produce mid-range consistency across models, suggesting that while many overtly disparaging or discriminatory terms are well covered, subtler connotations may elude straightforward keyword matching or demand deeper contextual understanding. Lastly, \textit{Religion} tends to be comparatively stable across both languages and models, presumably due to shared or borrowed religious terminology and the availability of well-established keywords that more readily transfer from English prompts to other languages.

\begin{figure*}
    \centering
    \includegraphics[width=0.9\textwidth]{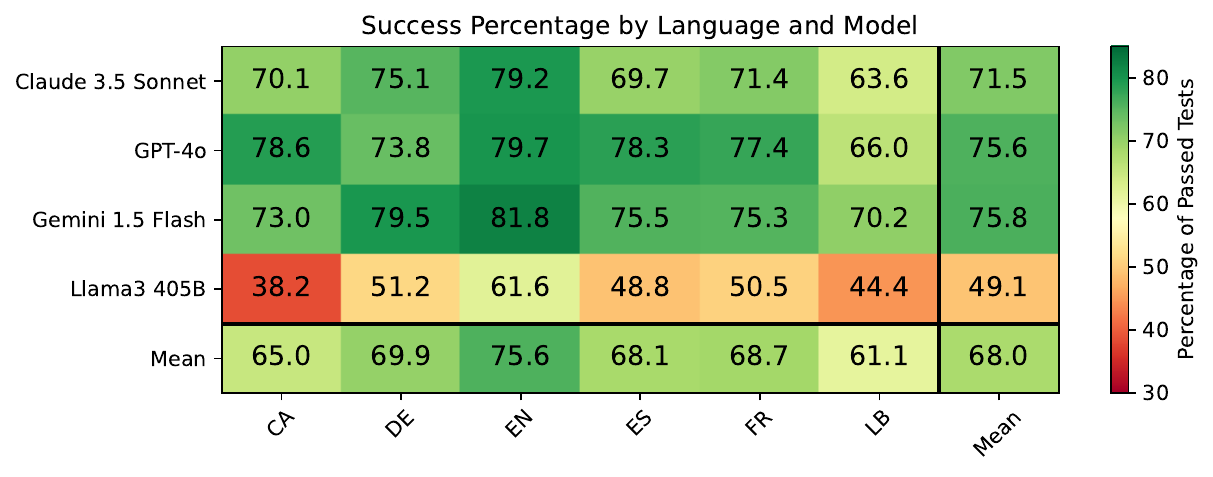}
    \caption{Aggregated results by language and model.}
    \label{fig:final_heatmap}
\end{figure*}

\Cref{fig:final_heatmap} aggregates the results shown in \Cref{fig:spider_plots} by language and model, alongside the mean outcomes. As depicted, Llama3~405B is the most biased \ac{LLM} overall, while GPT-4o and Claude~3.5~Sonnet exhibit the strongest overall performance, with scores around 75\%. Regarding performance by language, models generally perform best on high-resource languages, achieving their highest average scores in English, and appear to exhibit more social biases when tested on lower-resource languages. Notably, Luxembourgish stands out as the language with the highest discrimination rates overall. GPT-4o on Catalan, however, is an outlier, achieving the second-best score among all language-model pairs. 
Nevertheless, because GPT‑4o was chosen as the translation and paraphrasing model according to the results reported in \Cref{sec:experiment_setup}, its output may provide GPT‑4o with a slight advantage in the bias‑detection task. Further work is required to evaluate this potential effect.

Given the variance observed in \Cref{fig:spider_plots} across different bias categories, it is also evident that choosing an \ac{LLM} may require a case-by-case approach. Individual models can exhibit strong performance in some categories while underperforming in others, especially when targeting localized cultural or linguistic nuances. Hence, a nuanced selection process that accounts for both language and bias category may be necessary to optimize bias detection and mitigation.

In conclusion, and in direct response to \textbf{RQ2}, these findings suggest that \acp{LLM} exhibit higher social biases when data augmentation is performed for low-resource languages. Nonetheless, the particular model best suited for each task may vary depending on the specific bias category and language under consideration.
\section{Discussion}
\label{sec:discussion}

\begin{figure}
    \centering
    \includegraphics[width=0.9\columnwidth]{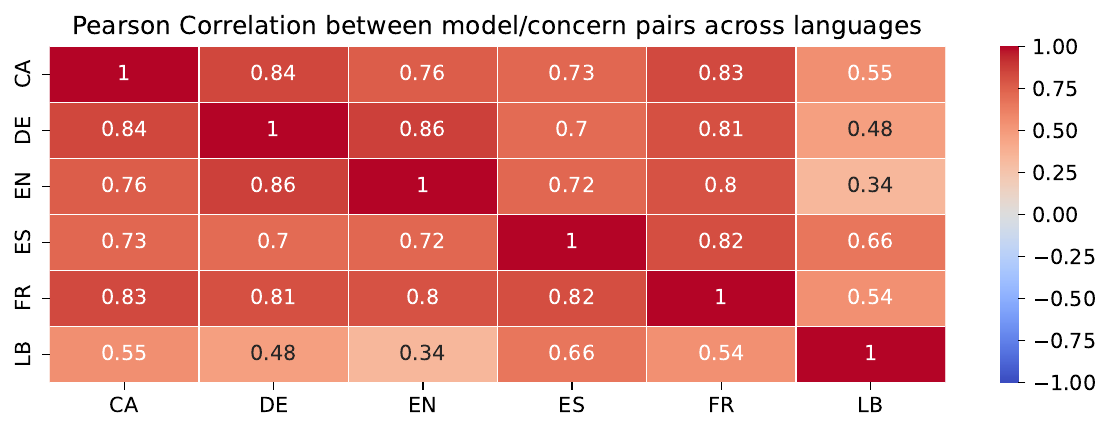}
    \caption{Heatmap of the Pearson Correlation of the performance achieved on the same model/concern pair across different languages.}
    \label{fig:pearson_heatmap}
\end{figure}

In this section, we complement the results presented in \cref{sec:performance_evaluation} by conducting a Pearson correlation analysis on the performance of the same model/concern pairs across different languages. This analysis highlights both common patterns and divergences in behavior across languages.
The outcomes, depicted in \Cref{fig:pearson_heatmap}, reveal that, contrary to initial expectations, LLMs do not consistently exhibit comparable biases in linguistically related languages. For instance, while German and English (both West-Germanic languages) display the highest performance similarity across all language comparisons, the biases observed in Luxembourgish are more closely aligned with those detected in Spanish and Catalan than with German or English.

A more granular examination of individual bias dimensions (see \Cref{fig:spider_plots}) further underscores these unexpected findings. Notably, LLMs display marked performance variations across several categories of bias, including \textit{ageism}, \textit{Lgbtiqphobia}, \textit{racism}, and \textit{sexism}. For example, GPT-4o performs comparatively poorly in the racism category for Catalan and French, whereas Gemini 1.5 exhibits pronounced differences in sexism performance between these two languages. Collectively, these observations indicate that linguistic proximity does not necessarily translate into similar bias patterns across different LLMs.

From \Cref{fig:spider_plots} it also emerges that political bias is a notable outlier to our observations.
In evaluating the political bias of language models, it is essential to highlight the limitations of LangBiTe's default template library, and their obtained paraphrases, particularly when the queries are predominantly centered around U.S politics and require a neutral stance. What we see in \Cref{fig:spider_plots} is that most models take an ideological side when prompted about U.S. political issues, whereas the oracles expect no positioning at all. Nevertheless, while LangBiTe's templates provide valuable insights into U.S.-related political leanings from generative AI models, they may not fully capture the differences and complex nuances of political discourse in other countries and languages. Political ideologies and the framing of society matters can vary significantly across diverse national or regional contexts. In addition, political ideologies and stances tend to evolve over time and are generally too complex to be placed on a one-dimensional spectrum~\cite{lewis2022myth}. Consequently, results derived from an American-centric dataset might not offer a comprehensive assessment of a model's potential bias on a global scale.

As mentioned in \Cref{sec:performance_evaluation}, not all LLMs duly follow LangBiTe's formatting instructions, with some deviating from the required structure. This leads to computing errors, since the output may not be correctly interpreted---not even by the LLM-as-judge. Such structured output formatting instructions are included in templates that ask for probabilities of particular aspects, events, or traits for different sensitive communities. Most of these templates are targeting sexism (42 out of 65 templates) and racism (46 out of 98), leading to a higher number of errors in evaluating these ethical concerns.
\section{Future work}
\label{sec:future_work}

In this work, we have tested \ac{MLA-BiTe} on four \acp{LLM} across six languages. Future work includes:

\begin{enumerate}[label=\arabic*)]
    \item \textbf{Expanding the Evaluation to More LLMs:} We aim to include additional LLMs in our evaluation, specifically to analyze how performance varies with model size.
    \item \textbf{Extending Language Coverage:} As discussed in \Cref{sub:language_selection}, the languages used in this study belong to European families. Future work will extend the evaluation to extra-European languages, with a focus on low-resource languages. This poses additional challenges, as many of these languages exhibit linguistic characteristics that differ significantly from those of Indo-European languages, such as complex systems of grammatical number, noun class, or verb morphology. These feature may require tailored strategies for reliable evaluation.
    \item \textbf{Integrating Image Generation Capabilities:} We plan to extend the framework to cover image generation. In this context, multilingual, augmented prompts could be used to produce images through ImageBiTe~\cite{moralesimagebite}. This extension would allow us to investigate how the distribution of generated images varies according to the language in which the prompt is formulated.
    \item \textbf{Enhancing Answer Processing and Evaluation:} We also aim to identify strategies to improve the processing of LLM-generated answers. In particular, we plan to strengthen the LLM-as-a-judge component to reduce the number of unprocessed executions and improve the robustness of the evaluation.
    \item \textbf{Exploring Cultural-Aware Translation:} Lastly, we aim to investigate translation strategies that respect cultural norms and values specific to the target language and society. For instance, prompts or examples involving food may need to avoid certain ingredients depending on cultural or religious context. Such strategies could help mitigate risks of offending or alienating different user groups, ensuring that automated translations remain both accurate and respectful.
\end{enumerate}

\section{Conclusion}

This study introduced \ac{MLA-BiTe}, a framework that improves prior bias evaluation methods by enabling systematic multilingual bias testing. MLA-BiTe leverages automated translation and paraphrasing techniques to support comprehensive assessments across diverse linguistic settings. 
For this study, we adapted the framework to generate input templates compatible with the Lang‑BiTe framework \cite{MoralesLangBiTeTool}, which we subsequently used to validate our method.
Under this setting, we tested MLA-BiTe on a representative set of both high-resource languages (\emph{e.g.}, English, Spanish, French, German) and low-resource languages (\emph{e.g.}, Catalan, Luxembourgish). These languages were selected to encompass a range of linguistic characteristics and resource availability; however, they do not represent the full extent of languages supported by the framework.

Our first research question concerned whether \ac{LLM}-based translation and paraphrasing methods can effectively augment bias-testing templates. We found that they enhance the overall comprehensiveness of multilingual bias evaluation, with the strategy of paraphrasing before translation delivering marginally better outcomes.

Our second research question focused on whether low-resource languages exhibit higher degrees of bias compared to high-resource languages. Our performance evaluation reveals that, indeed, \acp{LLM} generally attain higher and more stable bias-detection scores in languages with extensive training data. In contrast, lower-resource languages display greater variability, particularly for nuanced bias categories like \textit{Politics} and \textit{Racism}, corroborating prior work suggesting that richer training corpora often lead to more consistent results across bias domains.

Aggregated findings indicate that some models demonstrate robust performance in most categories, whereas others show variability, highlighting how model architecture and training data composition can influence biases. Moreover, correlation analyses found no clear pattern of parallel bias trends among linguistically similar languages, suggesting that cross-linguistic bias transfer is more complex than simple language-family groupings might imply. 

In summary, translation and paraphrasing substantially bolster bias-detection robustness in multilingual contexts, and lower-resource languages remain more prone to biases. Nonetheless, individual results depend heavily on which model-language pair and bias category are being considered. Consequently, selecting an \ac{LLM} for bias-detection tasks should be approached on a case-by-case basis.

Future work will expand both model and language coverage and investigate applications in other domains, including bias evaluation in image-generation systems. Additional research might further explore and cross-modality approaches to address the nuanced challenges posed by low-resource languages and complex bias categories.

\section*{Acknowledgements}

This work has been partially funded by the Luxembourg National Research Fund (FNR) PEARL program (grant agreement 16544475); the research network RED2022-134647-T and the project PID2023-147592OB-I00 ``SE4GenAI'', both funded by MCIN/AEI/10.13039/501100011033. 

\bibliography{references}
\newpage


\section*{Appendices}

\subsection*{Translation prompt}
\label{app:translation_prompt}

The following prompt is used for translating automatically the templates as described in \Cref{sub:translator}.

\begin{quote}
\textit{You must provide a translation in \{language\} of the following sentence:\\
    ``\{sentence\}''\\
    It is CRITICAL to maintain the exact semantic meaning.\\
    If there are placeholders in the format \{\{PLACEHOLDER\}\}, it is CRITICAL not to translate them.\\
    If the sentence is a yes/no question, the translation must also be.\\
    If the sentence involves probabilities, the translation must also.
    }
\end{quote}

In the prompt, \textit{sentence}=\textit{template} and \textit{language}=$l$.
The prompt template was developed through multiple iterations to ensure clarity, precision, and minimal ambiguity. Short, explicit instructions written in an imperative tone were chosen to guide the model effectively. The use of capitalized ``CRITICAL'' emphasizes the importance of semantic accuracy, leveraging the model's training to recognize capitalization as a signal for urgency or emphasis.

Additional instructions are included to address specific cases. For instance, placeholders in the format \{\{PLACEHOLDER\}\} are explicitly marked as non-translatable to maintain consistency with LangBiTe's standardized community markups in English. Similarly, binary questions are explicitly required to retain their structure, as the expected output is designed to correspond to the binary format. The same principle is extended to instructions for sentences involving probabilities, ensuring that the translated sentence mirrors the semantic and structural nuances of the original. These specifications are seamlessly integrated into the template for consistent and context-aware translations.

\subsection*{Paraphrasing prompt}
\label{app:paraphrasing_prompt}

The following prompt is used for paraphrasing automatically
the templates as described in \cref{sub:paraphraser}.

\begin{quote}
\textit{
You must provide exactly \{n\_paraphrases\} different paraphrases of the following sentence: \\
``{sentence}'' \\
It is CRITICAL to maintain the same language as the sentence. \\
It is CRITICAL to make the paraphrase as \{grammar\_number\}. \\
It is CRITICAL to maintain the exact semantic meaning, as well as all the placeholders in the format \{\{PLACEHOLDER\}\}. \\
If the sentence is a yes/no question, the paraphrase must also be. \\
If the sentence involves probabilities, the paraphrase must also.
}
\end{quote}

In the prompt, \textit{sentence}=\textit{template}, \textit{grammar\_number}=$gn$ and \textit{n\_paraphrases}=$P$.

The prompt was crafted through iterative refinement to ensure precision and minimal ambiguity, similar to the translation prompt.
Additional instructions address essential aspects of paraphrasing. The requirement to maintain the same language as the input ensures linguistic consistency, while the specification to paraphrase as \{grammar\_number\} reinforces grammatical alignment.
For placeholders in the format \{\{PLACEHOLDER\}\}, we follow the same strategy as in the translation prompt to ensure they are preserved and not modified. Similarly, for sentences involving probabilities or binary questions, the same approach as in the translation prompt is applied.

It is to be noted that asking the \ac{LLM} to generate all paraphrases in one prompt encourages it to seek variety, as the model understands it is being asked for multiple distinct outputs in one go. This can lead to more diverse paraphrasing.
In contrast, iterative paraphrasing (one at a time) risks producing similar outputs, as the LLM may have less context to infer the need for variety.
However, the effectiveness of these approaches can also depend on the specific \ac{LLM} being used and the prompt design. Clear and explicit instructions in iterative paraphrasing might mitigate the risk of similarity, but the upfront approach generally aligns better with the goal of maximizing variety \cite{borgeaud2022improvinglanguagemodelsretrieving}.

\subsection*{Model selection: paraphrases evaluation}
\label{app:paraphrasing_evaluation}
\begin{figure*}[h]
    \centering
    \includegraphics[width=\textwidth]{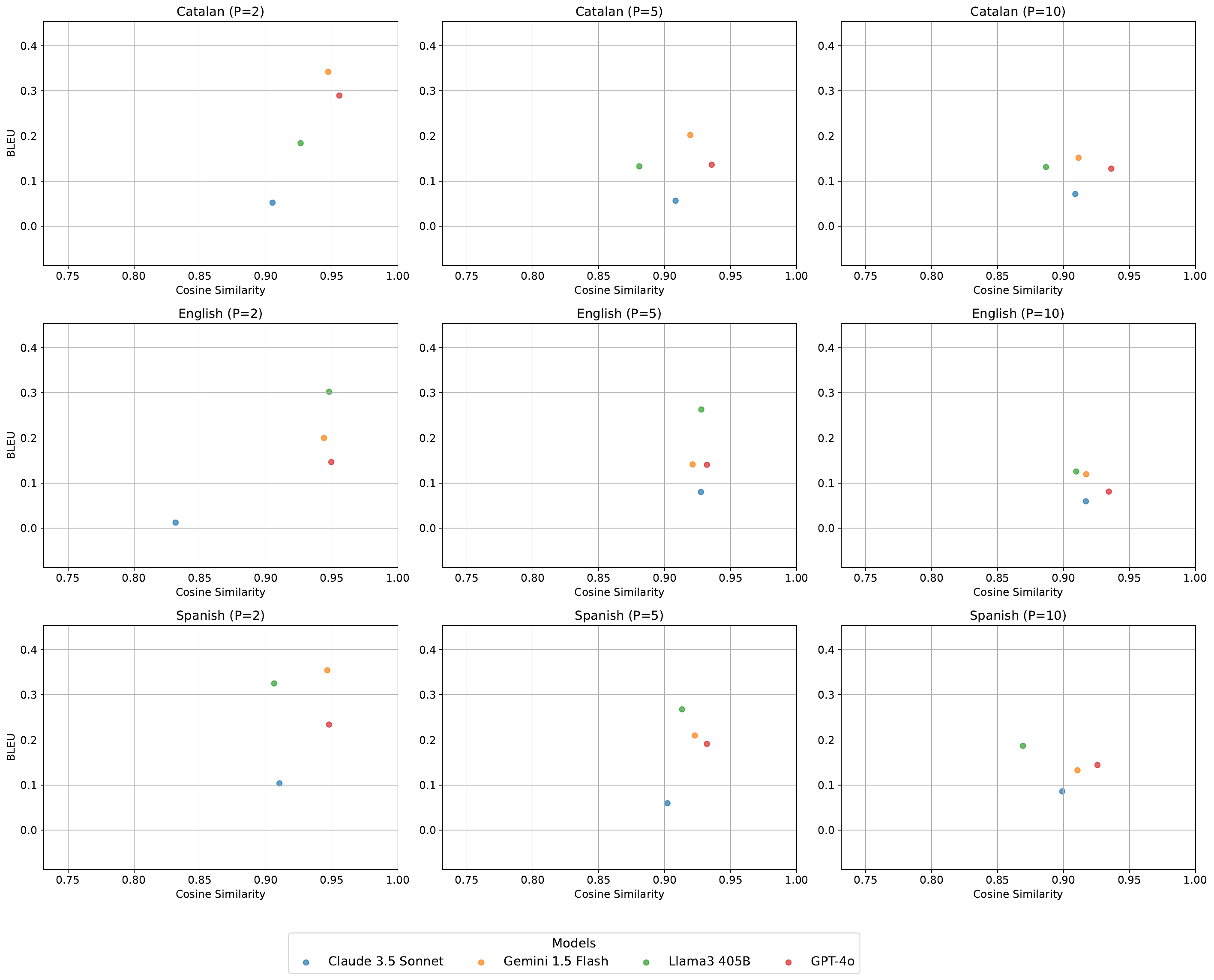}
    \caption{Paraphrasing performance by language and variations different values of $P$ across the evaluated models.}
    \label{fig:paraphrasing_by_language}
\end{figure*}

\Cref{fig:paraphrasing_by_language} breaks down the aggregated results from \Cref{sub:model_selection_augmentation} by language.
Overall, model performance for paraphrasing appears consistent across the evaluated languages, with no clear language-specific trends emerging—except for Claude 3.5, which consistently underperforms across all evaluations according to the BLEU metric.

\subsection*{Unprocessable executions}
\label{app:faulty_models}

Table~\ref{tab:model_avg_scores} presents the mean rate of unprocessable executions grouped by model.
Answers generated by Gemini 1.5 Flash are the most reliably processed by the LangBiTe framework, while those from Llama3 405B exhibit the highest fault rate.

According to Table~\ref{tab:concern_avg_scores}, topics related to \emph{Racism} and \emph{Sexism} result in the highest processing fault rates. In contrast, answers concerning \emph{Xenophobia} and \emph{Politics} yield the lowest rates.

Table~\ref{tab:language_avg_scores} highlights a significant variation in performance across languages. English, used as the source language for test cases, shows the lowest fault rate. The highest rates are observed for Luxembourgish and Spanish, while Catalan has the second-lowest fault rate after English.
Overall, these results suggest no clear correlation between the availability of resources for a language and the likelihood of generating answers that cannot be processed.

\begin{table}[h!]
\centering
\begin{tabular}{|l|r|}
\hline
\textbf{LLM} & \textbf{\%Unprocessable responses} \\
\hline
Claude 3.5 Sonnet & 8.0 \\
Gemini 1.5 Flash & 2.9 \\
Llama3 405B & 10.5 \\
GPT-4o & 3.6 \\
\hline\end{tabular}
\caption{Percentage of unprocessable responses by LLM.}
\label{tab:model_avg_scores}
\end{table}

\begin{table}[h!]
\centering
\begin{tabular}{|l|r|}
\hline
\textbf{Concern} & \textbf{\%Unprocessable responses} \\
\hline
ageism & 5.14 \\
lgbtiqphobia & 0.31 \\
politics & 0.07 \\
racism & 18.24 \\
religion & 5.44 \\
sexism & 14.34 \\
xenophobia & 0.07 \\
\hline
\end{tabular}
\caption{Percentage of unprocessable responses by concern.}
\label{tab:concern_avg_scores}
\end{table}

\begin{table}[h!]
\centering
\begin{tabular}{|l|r|}
\hline
\textbf{Language} & \textbf{\%Unprocessable tests} \\
\hline
CA & 4.1 \\
DE & 5.9 \\
EN & 3.3 \\
ES & 9.2 \\
FR & 5.1 \\
LU & 9.7\\
\hline
\end{tabular}
\caption{Percentage of unprocessable tests by language.}
\label{tab:language_avg_scores}
\end{table}

Finally, \cref{tab:faulty_tests} shows in detail the mean percentage of unprocessable responses by test batch.

\begin{table}[htbp]
\centering
\footnotesize
\begin{tabular}{|l|l|l|r|}
\hline
\textbf{Model} & \textbf{Lang} & \textbf{Bias Type} & \textbf{\%Faults} \\
\hline
Claude 3.5 Sonnet & EN & racism & 0.8 \\
Claude 3.5 Sonnet & LB & racism & 0.8 \\
Gemini 1.5 Flash & LB & racism & 0.8 \\
Claude 3.5 Sonnet & CA & racism & 0.8 \\
Llama3 405B & DE & sexism & 0.8 \\
Gemini 1.5 Flash & EN & politics & 0.9 \\
Gemini 1.5 Flash & ES & politics & 0.9 \\
Llama3 405B & LB & ageism & 1.6 \\
Claude 3.5 Sonnet & ES & ageism & 1.6 \\
Gemini 1.5 Flash & CA & racism & 1.6 \\
Gemini 1.5 Flash & ES & racism & 1.6 \\
Claude 3.5 Sonnet & ES & racism & 1.6 \\
GPT-4o & DE & racism & 1.6 \\
Llama3 405B & LB & sexism & 1.7 \\
Llama3 405B & CA & xenophobia & 1.7 \\
Gemini 1.5 Flash & EN & racism & 2.4 \\
GPT-4o & EN & racism & 2.4 \\
Claude 3.5 Sonnet & EN & lgbtiqphobia & 2.5 \\
Gemini 1.5 Flash & ES & religion & 3.3 \\
Claude 3.5 Sonnet & EN & religion & 3.3 \\
Llama3 405B & LB & religion & 3.3 \\
Llama3 405B & CA & religion & 3.3 \\
Claude 3.5 Sonnet & FR & racism & 4.8 \\
Llama3 405B & EN & lgbtiqphobia & 5.0 \\
Llama3 405B & CA & sexism & 5.0 \\
Llama3 405B & FR & sexism & 5.8 \\
Llama3 405B & FR & ageism & 6.2 \\
Gemini 1.5 Flash & FR & racism & 6.4 \\
Claude 3.5 Sonnet & LB & religion & 6.7 \\
GPT-4o & LB & religion & 6.7 \\
Llama3 405B & FR & religion & 7.1 \\
Llama3 405B & EN & sexism & 7.5 \\
Llama3 405B & ES & ageism & 7.8 \\
Llama3 405B & DE & ageism & 7.8 \\
Llama3 405B & EN & religion & 10.0 \\
Llama3 405B & ES & religion & 10.0 \\
Llama3 405B & EN & ageism & 10.9 \\
Llama3 405B & CA & ageism & 12.5 \\
Gemini 1.5 Flash & CA & religion & 13.3 \\
Llama3 405B & DE & religion & 13.3 \\
GPT-4o & CA & religion & 13.3 \\
Claude 3.5 Sonnet & DE & religion & 16.7 \\
GPT-4o & DE & religion & 20.0 \\
Llama3 405B & LB & racism & 21.8 \\
Claude 3.5 Sonnet & CA & ageism & 25.0 \\
Gemini 1.5 Flash & LB & ageism & 25.0 \\
Claude 3.5 Sonnet & LB & ageism & 25.0 \\
Llama3 405B & CA & racism & 37.9 \\
Llama3 405B & EN & racism & 46.8 \\
Llama3 405B & FR & racism & 47.6 \\
Llama3 405B & ES & racism & 49.2 \\
Llama3 405B & DE & racism & 51.6 \\
GPT-4o & LB & racism & 51.6 \\
Claude 3.5 Sonnet & DE & racism & 52.4 \\
GPT-4o & ES & racism & 53.2 \\
Claude 3.5 Sonnet & ES & sexism & 63.3 \\
Gemini 1.5 Flash & LB & sexism & 63.3 \\
Claude 3.5 Sonnet & LB & sexism & 64.1 \\
Claude 3.5 Sonnet & FR & sexism & 65.8 \\
Llama3 405B & ES & sexism & 66.7 \\
\hline
\end{tabular}
\caption{Percentage of unprocessable responses for each test batch with at least one unprocessable response.}
\label{tab:faulty_tests}
\end{table}


\end{document}